\def\year{2020}\relax
\documentclass[letterpaper]{article} 
\usepackage{aaai20}  
\usepackage{times}  
\usepackage{helvet} 
\usepackage{courier}  
\usepackage[hyphens]{url}  
\usepackage{graphicx} 
\usepackage{pifont}
\newcommand{\cmark}{\ding{51}}%
\newcommand{\xmark}{\ding{55}}%
\newcommand{\crossedcmark}{\textcolor{black}{\ding{51}}{\small\textcolor{black}{\kern-0.7em\ding{55}}}}
\usepackage{adjustbox}
\usepackage{lipsum}
\usepackage{tabularx}
\newcolumntype{P}[1]{>{\centering\arraybackslash}p{#1}}
\usepackage{subcaption}
\usepackage[show]{notes}
\newcommand{\ignore}[1]{}
\newcommand{\com}[1]{}

\newcommand{\rev}[1]{\textcolor{black}{\rm #1}}
\usepackage{float}
\restylefloat{table}
\usepackage[table,xcdraw]{xcolor}
\usepackage{graphicx}  
\title{It's a Thin Line Between Love and Hate: \\
Using the Echo in Modeling Dynamics of Racist Online Communities}

\author{Eyal Arviv,  Simo Hanouna, Oren Tsur \\
Software and Information Systems Engineering\\
Ben Gurion University of the Negev\\
\{eyalar,hanouns\}@post.bgu.ac.il, orentsur@bgu.ac.il}
 
\begin{document}

\maketitle

\begin{abstract}
The \texttt{(((echo)))} symbol -- triple parenthesis surrounding a name, made it to mainstream social networks in early 2016, with the intensification of the U.S. Presidential race. It was used by members of the alt-right, white supremacists and internet trolls to tag people of Jewish heritage -- a modern incarnation of the infamous yellow badge (Judenstern) used in Nazi-Germany. Tracking this trending meme, its meaning, and its function has proved elusive for its semantic ambiguity (e.g., a symbol for a virtual hug). 

In this paper we report of the construction of an appropriate dataset allowing the reconstruction of networks of racist communities and the way they are embedded in the broader community. We combine natural language processing and structural network analysis to study communities promoting hate. In order to overcome dog-whistling and linguistic ambiguity, we propose a multi-modal neural architecture based on a BERT transformer and a BiLSTM network on the tweet level, while also taking into account the users ego-network and meta features. Our multi-modal neural architecture outperforms a set of strong baselines.
We further show how the the use of language and network structure in tandem allows the detection of the leaders of the hate communities. \rev{We further study the ``intersectionality'' of hate and show that the antisemitic echo correlates with hate speech that targets other minority and protected groups}. Finally, we analyze the role IRA trolls assumed in this network as part of the Russian interference campaign. Our findings allow a better understanding of recent manifestations of racism and the dynamics that facilitate it. 
\end{abstract}

\section{Introduction}
\label{sec:intro}
Hate speech proliferates in social media \cite{waseem2016hateful,laub2019hateGlobalComparison}. 
While harassment may be targeted at any individual, hate speech typically references groups and targets individuals for their group identity. \rev{Women, people of color, the LGBT community, Muslims, immigrants, and Jews are among the most targeted groups}. Recent studies report on a surge in Islamophobia \cite{sunar2017long,akbarzadeh2016muslim,osman2017retraction}, Antisemitism \cite{adl2020antisemitism,zannettou2020quantitative}, xenophobia \cite{iwama2018understanding,entorf2019refugees} and hate toward other groups \cite{levin2018report,dodd2017anti,edwards2018effect,perry2020planting}.

Online hate speech is not merely an online inconvenience. It directly manifests itself in ``real life'' through shooting, bombing, stabbing, beating, and vandalizing. These violence incidents are often linked directly to online activity \cite{splc2018altright,altrightpipeline,violence2019,8chan}. 
A recent U.N. report on the Freedom of Religious and Belief, transmitted by the Secretary General amidst the global rise in antisemitism, asserts that ``\emph{antisemitism, if left unchecked by Governments, poses risks not only to Jews, but also to members of other minority communities. Antisemitism is toxic to democracy}'' \cite{shaheed2019unreport}.

\rev{Like misery, racial hate likes company. Hate is not expressed and promoted by random individuals -- rather, it is the product of \emph{communities}, often embedded in larger communities. Therefore, in order to better understand the social mechanisms involved in the promotion of online hate, we need to understand how these communities are structurally organized and how they use specific, often elusive, language to promote their cause.}

In this paper we combine structural analysis of Twitter networks with textual analysis (NLP) to identify hate communities that are embedded in broader networks. Using the \emph{echo}, an elusive antisemitic meme, as a starting point, we demonstrate how networks promoting hate can be recovered, which in turn enables us to identify key features that contribute to the promotion of hate.

\rev{Specifically, we address the following questions:\begin{itemize}
\item {\bf Disambiguation} -- can we properly identify hate speech and disambiguate the uses of nuanced language and memes that are used both legitimately and in a dog-whistling manner?
\item {\bf Text and social structure} -- can we leverage the network structure in order to achieve detection of hate-mongers?
\item {\bf Intersectionality of hate} -- how is the antisemitic echo meme related to other forms of hate speech and other minority groups?
\item{\bf Linguistic variation} -- can the evolution of the echo meme be interpreted according to linguistic theory?
\end{itemize}}

The two former questions are addressed algorithmically in both unsupervised and minimally supervised way, as we propose a multi-modal neural architecture based on a BERT Transformer and a BiLSTM for the utterance (tweet) level that feeds into another classifier that also models the user ego-network and metadata. The two latter questions, are addressed qualitatively.


\begin{table*}[ht!]
\centering
\begin{tabular}{c|c|m{15cm}}
\hline
\rowcolor[HTML]{C0C0C0} 
& LABEL & TWEET \\ \hline\hline
1 & HM & \texttt{Don't Trigger Mr Trump (((Rosengerg))) it might cause him to fire up the ovens \#OvenWorthy} \\ \hline
2 & HM & \texttt{RT @USR: Andrew Breitbart was murdered by (((Globalists))). \#PizzaGate }    \\ \hline
3 & HM & \texttt{Trybalist symbol for establishment climbers? Most of our critiques have 3 (((brackets around their names))) \& have black-rim glasses. Bizarre.}  \\ \hline
4 & HM & \texttt{That's because Trump doesn't hate white Gentiles like ((((((((((THEY))))))))))) do.}  \\ \hline \hline
5 & R & \texttt{ADL adds (((echo))) symbol to hate list}     \\ \hline
6 & R & \texttt{People are putting ((( echoes ))) around their names on Twitter - here's why}      \\ \hline
7 & R  & \texttt{@USR alright wise one... What does ((( ))) around someone's name?}    \\ \hline \hline
8 & N & \texttt{We're (((LIVE))) on the radio near you --\textgreater ~its \#LightOnLive with <NAME>, from now till 6am on \#Live919FM}      \\ \hline
9 & N & \texttt{@USR THIS WOMAN NEEDS A BIG HUG (((HUG)))}    \\ \hline
10 & N & \texttt{can u get any cooler than that ((((nope)))) } \\
\end{tabular}
\caption{Echo tweets and their type. HM: hate-mongering; R: response to HM. N: Neutral (not hate); User names and real names were replaced by @USR and NAME, respectively. A tweet containing expressive lengthening can be seen in the fourth example of this table.}
\label{tab:annotation_table}
\end{table*}

\section{Background and Related Work}
\label{sec:related}

\paragraph{The Echo} The triple parentheses, or triple brackets, also known as the \texttt{(((echo)))}, is an antisemitic symbol that is used to highlight the names of individuals of a Jewish background (e.g., Jeffery Goldberg, Editor-in-Chief of The Atlantic), organizations owned by Jewish people (e.g., Ben \& Jerry's), or organizations accused of promoting ``Jewish globalist values'' (e.g., the International Monetary Fund). Originally an audial meme used at the podcast {\em The Daily Shoah}, the meme was popularized in a textual form in the white-supremacy blog \textit{The Right Stuff}. The echo slowly drifted from fringe websites to mainstream social platforms like Twitter, reaching a wider audience and expanding its user base. 
Typical examples of an antisemitic use of the echo are presented in rows 1--4 in Table \ref{tab:annotation_table}. Tweets 1,2 were posted by regular users, referring to an individual (\#1), and promoting an antisemitic trope about Jewish domination (\#2). The 3rd tweet, promoting a similar antisemitic trope, was posted by a high profile organization -- the official \texttt{@WikiLeaks} account, after the organization was criticised for alleged ties to the Russian Intelligence. The tweet was removed within hours, not before being retweeted and stared hundreds of times.

Members of hate communities often use specific language and symbols to convey their affiliation and promote their agenda. Unique, vague and ambiguous patterns of language may arise from community culture and are often used as a dog-whistling practice used in order to avoid detection and suspension\footnote{While some social platform, e.g., Reddit, 4chan and Gab \cite{zannettou2018gab,lima2018inside} have limited or no moderation, platforms like Twitter officially prohibit hate speech.}. While used as a hate symbol by some users, the echo has multiple senses, e.g., `broadcasting', `emphasis' or a `virtual hug' (see Table \ref{tab:annotation_table} 8-10). In Section \ref{subsec:ideas_hashtags} we further discuss special lingo and ambiguous terms. 

\rev{The recent rise in online hate speech attracts significant body of research. Broadly speaking, this body of research could be broken to two main categories, focusing on two different perspectives: (i) the algorithmic detection of hate-speech, and (ii) social analysis of the use (and users) of hate speech. In the remainder of this section we provide brief survey of relevant work.}

\paragraph{Hate, Trolls and Online Culture} The {\em alt-right}, short for `the alternative right' is a term referring to a collection of organizations and individuals sharing extreme right-wing ideology that ranges from classic far-right ideology to open white-nationality and white-supremacy. While traditional Internet trolls are not promoting a specific ideology \cite{phillips2015we}, alt-right trolls, rooted in Internet culture, seek to promote an extreme political agenda \cite{hawley2017making}. The similarity between gamer-gate trolls and the online activity of members of the alt-right is explored in \cite{bezio2018ctrl}. 
 
Hate speech is especially habitual in Gab and some forums on 4chan and Reddit \cite{hine2017kek,nagle2017kill,zannettou2018gab,grover2019detecting}. These platforms support a community structure in an almost explicit way\footnote{4chan and Reddit communities are defined by the boards and subreddits they subscribe to (e.g., 4chan/pol and reddit/altright). While the design of Gab is similar to Twitter, it brands itself as the ``free speech'' platform, thus attracts users that are banned from other networks for promoting hate.} and users adopt specific language to signal their affiliation and further enhance community bonds \cite{tuters2019they,zannettou2020quantitative}. On Twitter, on the other hand, communities are formed implicitly, as individuals follow or engage with other (like minded) individuals, thus the habit of signaling affiliation through the use of specific language and memes is of increased significance. However, since Twitter is more tightly moderated than 4chan, Reddit or Gab, the use of language tend to be more nuanced. 

\paragraph{Detection of Hate Speech} The use of code words, ambiguity and dog-whistling pose significant challenges to text-based detection of hate-speech \cite{davidson2017automated,ribeiro2017like}.  
The detection of implicit forms of hate speech is addressed by \cite{gao2017recognizing}, and \cite{magu2017detecting} detects the use of hate code words (e.g., google, skype, bing and skittle for Black, Jews, Chinese, and  Muslims, respectively). 

The use of demographic features such as gender and location in the detection of hate speech is explored by \cite{waseem2016hateful}, and user meta features, e.g., account age, posts per day, number of followers/friends, are used by \cite{ribeiro2017like}.

Computational methods for the detection of hate speech and abusive language range from the classic machine learning approaches such as SVM and logistic regression \cite{davidson2017automated,waseem2016hateful,nobata2016abusive,magu2017detecting}, to neural architectures such as RNNs and CNNs \cite{gamback2017using,zhang2016hate,del2017hate,park2017one}, and BERT transformers \cite{mozafari2019bert,samghabadi2020aggression,salminen2020developing}. \rev{For comparative surveys of taxonomies of hate speech and abusive language, available datasets, and models see \cite{salminen2018anatomy}, \cite{chen2019use}, and \cite{wullach2020towards}}.

The diffusion of hate in Twitter and Gab is modeled by \cite{ribeiro2017like} and \cite{mathew2019spread}, respectively. \rev{These works are close to our work as they address the user level, taking into account user meta features and network structure. However, the user meta features and network features are fixed and the textual analysis is basic. In contrast, we are concerned with the classification task rather than explicitly modeling the diffusion process. We put emphasis on the text, combining a BERT Transformer and a Bi-LSTM to classify users, and boost the model confidence by taking into account a weighted score assigned to other users in their network.}
The work of \cite{zannettou2020quantitative} is similar to ours in the sense that it is tracking a specific antisemitic meme (`the Happy Merchant') and address the community aspect of the phenomena. However, not only that our computational approaches are radically different, they are mostly concerned with quantifying specific antisemitic trends in Gab and 4chan/pol, while our focus is on disambiguation and user classification, and the analyzis of the memetics of the echo from a linguistic perspective. 

\rev{Our work differs from each of the works mentioned above in at least two of the following fundamental aspects: (i) we aim at detecting hate-monger and self organized hate communities, not only hateful posts, (ii) we harness both language (beyond keywords) and network structure in a multi-modal neural network, (iii) our dataset is radically different and significantly bigger than other datasets of hate and abuse in a mainstream platforms like Twitter, and (iv) our dataset was collected in an organic way by bootstrapping the ambiguous and elusive echo symbol. As such, it contains tweets posted by a diverse set of users, many of whom are not hate mongers, although they may use similar linguistic forms.\\
To the best of our knowledge, no previous work inherently combine the linguistic aspect and the network structure in a computational way, addressing the four questions listed in Section \ref{sec:intro}.}


\section{Data}
\label{sec:data}

\subsection{Echo Corpus}
A large dataset of over 18,000,000 English tweets posted by $\sim$7K echo users was constructed in the following manner\footnote{Twitter Search API ignores special characters, thus querying for the echo was not feasible.}: 

\begin{enumerate}
    \item {\bf Base Corpus}  We have obtained access to a random sample of 10\% of all public tweets posted in May and June 2016 -- the peak use of the echo. 
    \item {\bf Raw Echo Corpus} Searching the base corpus, we extracted all tweets containing the echo symbol, resulting in 803,539 tweets posted by 418,624 users. Filtering out non-English Tweets and users who used the echo less than three times we were left with $\sim$7K users\footnote{The echo is found in tweets written in multiple languages, particularly in East-Asian languages of which the user based is known for heavy use of ascii art and kaomoji \cite{mcculloch2019because}.}. 
    \item {\bf Echo Corpus} We used Twitter API to obtain the most recent tweets (up to 3.2K) of each of the users remaining in the English list\footnote{The data was collected in December 2016, amidst reports on the trending `echo'.}. This process resulted in $\sim$18M tweets posted by 7,073 users. We note that some of the accounts we found using echo were already suspended or deleted at the time of collection, thus their tweets were not retrievable. 
\end{enumerate}

\rev{To the best of our knowledge, this is the first time this dataset is being analyzed computationally and on large scale}.

\subsection{Data Annotation}
\label{subsec:gold_standard}
We sampled a thousand users from the dataset, inspected their use of the echo, and manually assigned each user one of three labels: HM (Hate Monger), R (Response) for users discussing the hate symbol, and N (Neutral) for users using the symbol in non-hate contexts. Examples of tweets from each category are presented in Table \ref{tab:annotation_table}, and descriptive statistics of the users of the different categories are presented under GOLD USERS in Table \ref{tab:class_stats}.

\begin{table*}[]
    \centering
    \begin{tabular}{l|c|c|c|c||c|c}
    & \multicolumn{4}{c||}{GOLD USERS} & \multicolumn{2}{c}{PREDICTED} \\
         Label & HM & R & N & R+N & HM & R+N\\
         \hline
         \hline
      Total \#Users   & 170 & 55 & 775 & 830 & 1136 & 5927 \\
      Total \#Tweets   & 339K & 141K & 2M & 2.15M & 2.26M & 15.44M \\
      \hline
       Avg. \#Days Active & 999$\pm$783 & 1910$\pm$973 & 1558$\pm$853 & 1582$\pm$866 & 1080$\pm$894 & 1511$\pm$876 \\
    Avg. Tweets/day & 11$\pm$19 & 7$\pm$9 & 19$\pm$37 & 18$\pm$36 & 15$\pm$31 & 32$\pm$96 \\
    Avg. \#Friends & 674$\pm$1445 & 741$\pm$1103 & 972$\pm$2136 & 957$\pm$2084 & 783$\pm$1527 &
    1224$\pm$5527 \\
    Avg. \#Followers  & 1022$\pm$2619 & 1067$\pm$2070 & 1941$\pm$5991 & 1884$\pm$5817 & 3848$\pm$60925 & 4432$\pm$85490 \\
    \hline
    Avg. \%Replies & 37$\pm$24 & 27$\pm$21 & 27$\pm$23 & 27$\pm$23 & 41$\pm$26 & 26$\pm$23 \\
    Avg. \%Retweets  & 34$\pm$24 & 26$\pm$24 & 24$\pm$22 & 24$\pm$22 & 31$\pm$25 & 24$\pm$22 \\
    Avg. \%URL  & 73$\pm$21 & 58$\pm$28 & 57$\pm$26 & 57$\pm$26 & 74$\pm$21 & 56$\pm$26 \\
    Avg. \%Hashtags & 16$\pm$14 & 22$\pm$23 & 20$\pm$23 & 20$\pm$23 & 18$\pm$16 & 19$\pm$24 \\
    \end{tabular}
    \caption{Account statistics derived from the annotated data (left) and predicted classes (right). Standard deviation is marked with $\pm$. Average days accounts are active, tweets per day, friends and followers are based on available account meta data. Average replies, retweets, URLs and hashtags ratios are based on tweets in the Echo corpus.}
    \label{tab:class_stats}
\end{table*}

\subsection{Network Statistics}
\begin{figure}[h!]
\centering
\includegraphics[width=.7\linewidth]{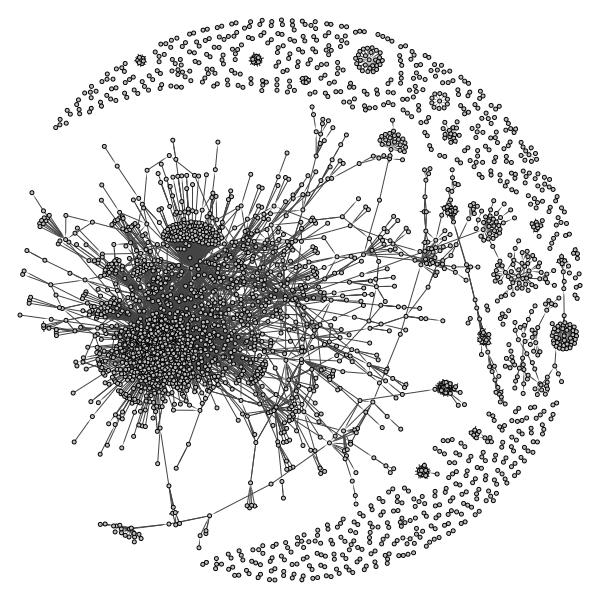}
\caption{Mentions network of echo users. Layout: force-directed. Minimum edge weight: 3.}
\label{fig:mention_raw}
\end{figure}

\begin{figure*}[h]
\begin{subfigure}[b]{.25\paperwidth}
\centering
\includegraphics[width=.25\paperwidth]{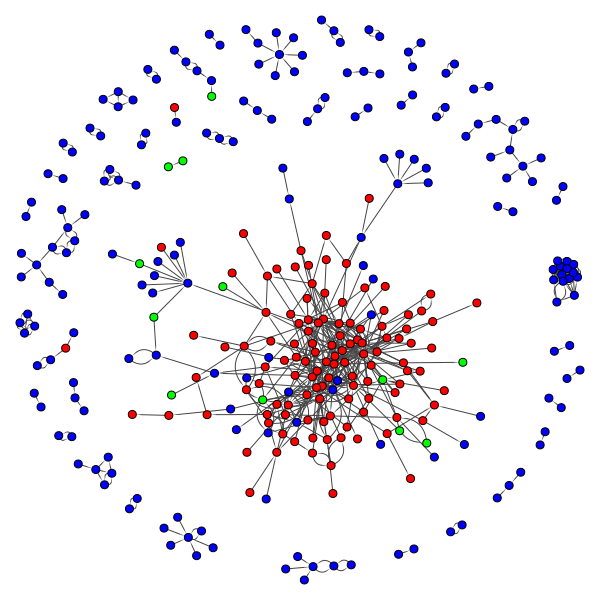}
\caption{Gold labels (no singletons)}
\label{subfig:mention_gold}
\end{subfigure}\hfill
\begin{subfigure}[b]{.25\paperwidth}
\centering
\includegraphics[width=.25\paperwidth]{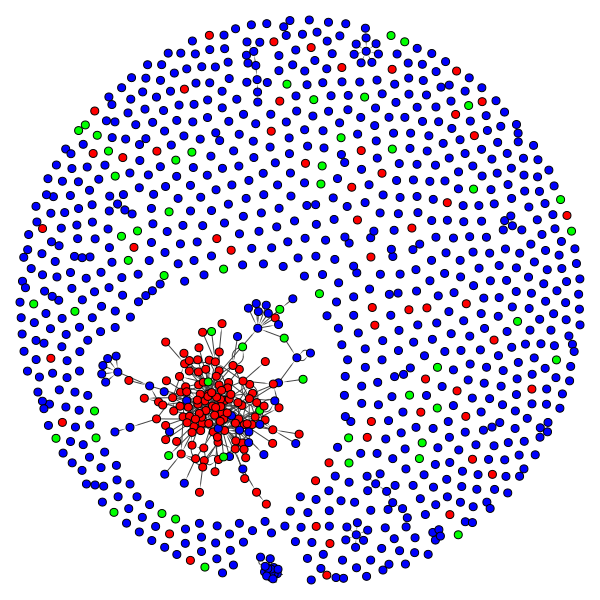}
\caption{Gold labels (with singletons)}
\label{subfig:mention_gold_sing}
\end{subfigure}\hfill
\begin{subfigure}[b]{.25\paperwidth}
\centering
\includegraphics[width=.25\paperwidth]{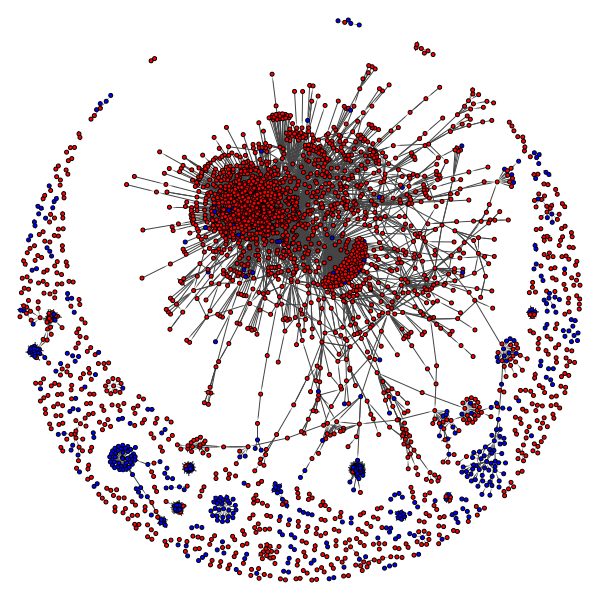}
\caption{Predicted labels (no singletons, R+N)}
\label{subfig:mention_pred}
\end{subfigure}\hfill
\caption{Mention based networks of echo users. Nodes colored by label -- HM in red, R in green, and N in blue.   Figures \ref{subfig:mention_gold} and \ref{subfig:mention_gold_sing} contain only the annotated subset of nodes. Figure \ref{subfig:mention_pred} is the same network presented in Figure \ref{fig:mention_raw}, the R and N classes are collapsed.}
\label{fig:networks_color}
\end{figure*}

Hate does not propagate in a void. Reconstructing the network of echo users enables us to identify structures, roles and interfaces that facilitate the propagation of hate-speech. Assuming that different types of engagement reflect different types of relations, we consider three different network semantics: mention-based, reply-based and retweet-based. In order to reduce noise we consider an edge only if its weight is higher than some threshold $\delta \geq 3$. 
The mention-based\footnote{Retweet and reply network are significantly sparser, but exhibit a similar structure in terms of communities. In the remainder of the paper we report results based on this network.} network presented in Figure \ref{fig:mention_raw} contains 3977 singletons (not presented in the figure), 2226 connected components (269 weak, 1993 strong), and a total of 3,092 nodes and 12,622 edges. Figures \ref{subfig:mention_gold} and \ref{subfig:mention_gold_sing} present only the nodes annotated as part of the gold standard, each node is colored by its label. The tendency of hate users to form tight communities is evident by the dominant cluster of red nodes that form the largest connected component (LCC). A detailed comparison between network statistics  of the full network and the LCC can be found in Table \ref{table:net_features} (top two rows).

\begin{table*}[h!]
\centering
\begin{tabular}{c|c|c|c|c|c|c|c|c}
Graph & \#Nodes & \#Edges & Density & Diameter & \#Triangles & Max \#triangles & \#Strong CC & \#Weak CC \\
\hline \hline
Full & 3092 & 12622 & 0.0013 & 20 & 24261 & 1988 & 1993 & 269 \\
LCC & 553 & 5215 & 0.0171 & 19 & 11114 & 1352 & n/a & n/a \\
HM & 730 & 5783 & 0.0109 & 11 & 11188 & 1728 & 387 & 34 \\
R+N & 2362 & 5018 & 0.0009 & 24 & 8918 & 669 & 1710 & 362 \\
\end{tabular}
\caption{Network (without singletons) features computed on the full mention network of echo users, and on its largest connected component (LCC), HM, and R+N subnetworks. We report on the following features for each network: Number of nodes, number of edges, density (computed without loops), diameter (within connected components), number of triangles, maximum number of triangles for a single node, and number of strongly and weakly connected components. Minimum edge weight: 3.}
\label{table:net_features}
\end{table*}

\begin{figure}[h!]
\centering
  \includegraphics[height=12cm,width=0.7\linewidth]{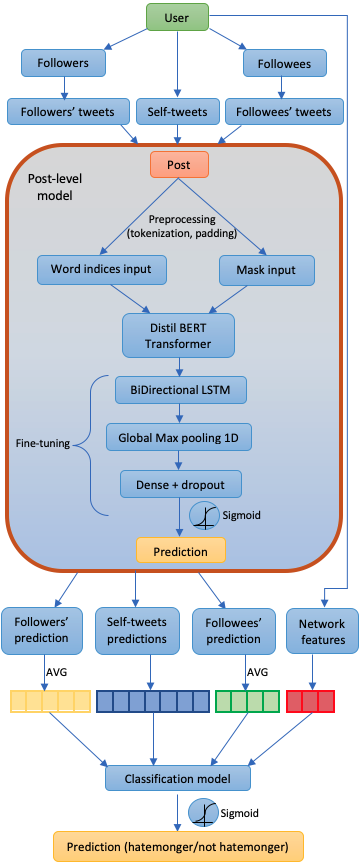}
  \caption{Multi-Modal neural architecture. The neural model process texts in different streams and treatments based on the structure of the social network.}
  \label{fig:neural_model}
\end{figure}

\section{Unsupervised Detection of Racist Users}
\label{sec:unsupervised}
While the analysis of the properties of the social network may shed light on the emergence of racist communities, some patterns may be missed due to data sparsity and the constraints imposed on data collection. We therefore opt for unsupervised content-based methods in order to discover disconnected individuals and clusters of like-minded racists. 

\paragraph{Experimental Setting} In order to achieve an abstract representation of topics and semantics we represented the text in two ways: word embeddings \cite{bojanowski2017enriching} and topic models \cite{wallach2009lda}. All tweets by a user were concatenated to one long text and users were represented in three different ways: (i) an average of their embeddings (EMBD), (ii) their single most salient topic (TM\textsuperscript{S}), and (iii) the full topic distribution vector for each user (TM\textsuperscript{F}). Clustering is done with the classic k-mean algorithm, assuming two settings: (i) three clusters, corresponding to the HM, R and N classes and, (ii) two clusters, collapsing the R and N classes to a single class of $\neg$HM.

\begin{table}[h!]
\centering
\begin{tabular}{c|c|c}
MODEL & RI\textsuperscript{2} & RI\textsuperscript{3} \\
\hline \hline
EMBD  & 0.687 & 0.687\\
TM\textsuperscript{S} & 0.801  & 0.672 \\
TM\textsuperscript{F} & {\bf 0.811} & {\bf 0.743}\\
\end{tabular}
\caption{Rand Index of clusters vs. Gold Standard set. Clusters are computed based on user's textual embeddings (EMBD), most salient topic representation (TM\textsuperscript{S}), and full topic distribution (TM\textsuperscript{F}). RI computed for two (RI\textsuperscript{2}) and three (RI\textsuperscript{3}) class/cluster settings. Embedding dimension: 300. K: 30.}
\label{table:ri}
\end{table}

\paragraph{Clustering Results} The Rand Index \cite{rand1971objective} is used to evaluate cluster quality against the gold standard set. All methods and settings achieved decent clustering results (see Table \ref{table:ri}). Best results were obtained  using the full distribution of topics ($k=30$). Figure \ref{subfig:mention_pred} presents the user cluster assignments (color) in the full network (singletons removed). Both the Rand Index (RI) results and the graphic visualization suggest a strong correlation between the network structure and the language used. These results are in line with previous studies of hate-speech in other platforms such as Gab and 4chan \cite{ribeiro2017like,zannettou2020quantitative}.

\section{Multi-Modal Neural Architecture}
\label{sec:nn}
\rev{Given that hate speech does not propagate in a void (see the previous section and related work), we propose a multi-modal neural architecture that takes into account the text of a user, as well as the texts of other users in her ego network. The main motivation for this approach is that multiple weak signals from user $u$'s ``neighborhood'' can be used to fine-tune the signal produced by user $u$ herself. This approach is common in sociology and demographic polling \cite{johnston1974local,latane1981psychology,sampson1988local} and we expect that it will be especially beneficial in the cases in which obscure, vague or ambiguous language is used.}

\paragraph{Post-Level Module (PLM)} \rev{The basic unit for classification is a single post (tweet). We fine tune a BERT transformer \cite{devlin2018bert} on the annotated dataset. Fine tuning is done after adding a bi-directional-LSTM with global max pooling, a dense, and a dropout layers. The architecture of the post level module is illustrated in the orange box in the center of Figure \ref{fig:neural_model}.}

\paragraph{User Network Module}\rev{ The post level module is used to process three distinct streams of tweets: (i) tweets of the user we wish to classify (user $u$), (ii) tweets of the users following $u$, and (iii) tweets of the users $u$ is following. The full architecture is illustrated in Figure \ref{fig:neural_model}. In this work, a user $v$ that mentioned $u$ $\geq \delta=3$ times is considered to be a follower of $u$, however, directed relations can be defined by other engagement patterns. 
The outputs of each of the three streams are processed slightly differently. While the results of the PLM of the user in question ($u$) are concatenated, the PLM results of each of her followers and followees are averaged on the user level (separately), thus each follower or followee contributes a single value to the final vector. 
All the PLM outputs are concatenated to a single vector that is composed of the all PLM predictions for tweets of $u$ (blue vector), a concatenation of all averaged scores for each of the followers and followees (yellow and green vectors, respectively). This vector is further concatenated with a vector of network features (red vector) of user $u$, e.g., in-degree, out-degree, betweeness, number of triangles $u$ is part of etc. The concatenated vector is fed to any classification model. We experimented with a three-layer FFNN, Gradient Boosted Machine (GBM) algorithms, and Logistic Regression. }

\begin{table}[h!]
\centering
\begin{tabular}{c|c|c|c|c}
MODEL & Precision & Recall & F1 & AUC \\
\hline \hline
$U$ & 0.692 & 0.607 & 0.613 & 0.925\\
$U$+$\overrightarrow{FU}$ &0.659 & 0.607 & 0.633 & 0.935 \\
$U$+$\overrightarrow{UF}$+$\overrightarrow{FU}$+$N$ & 0.873 & 0.666 & 0.755 & 0.959 \\
$U$+$N$ & 0.822 & {\bf 0.725} & 0.77 & 0.958\\
$U$+$\overrightarrow{UF}$ & 0.898 & 0.686 & 0.777 & 0.955\\
$U$+$\overrightarrow{UF}$+$N$ & {\bf 0.923} & 0.705 & {\bf 0.8} & 0.959 \\
\end{tabular}
\caption{Ablation results achieved by the multi-modal neural network. $U$: tweets of a single user $u$; $\overrightarrow{UF}$: tweets of users followed by $u$; $\overrightarrow{FU}$: tweets of the followers of $u$; $N$: $u$'s network features.}
\label{tab:neural_results}
\end{table}

\paragraph{Multi-Modal Neural Results} Using our proposed multi-modal architecture we achieve an F-score of 0.8. Table \ref{tab:neural_results} presents ablation results achieved using different components of the multi-modal architecture. Best results where achieved in the setting that includes the user's tweets, the tweets of the users she follow and the network features ($U$+$\overrightarrow{UF}$+$N$). It is worth noting that both network structure features and texts of followees provide a significant improvement over the baseline. These results are inline with our hypothesis that hate-mongers operate as a group and with the network coloring  reported in the previous section.

Interestingly, accounting for tweets posted by the followers of $u$ harms the classification. This result suggests an interesting dynamic between hate-mongers and mainstream users --  while extremists tend to mostly engage with other extremists, some mainstream users refer and engage with extremists. The reasons for this ``assymetry'' are beyond the scope of this work, however we hypothesis that some mainstream users are referring to and engaging with extremists either out of curiosity or in an effort to point to the phenomena. This hypotheses will be tested in future work.  


\section{Analysis and Discussion}
\label{sec:discussion}
The result reported in Sections \ref{sec:unsupervised} and \ref{sec:nn} demonstrate the strong connection between the network topology and the language used in different subcommunities. Moreover, using a multi-modal neural architecture we demonstrated that processing texts, while taking the network structure into account improves results significantly, especially in the case of vague or ambiguous language. In the remainder of this paper we further discuss various aspects related to the use of language, the network structure and the activity of hate groups.   

\subsection{Network Structure and Predicted Labels}
\label{subsec:structure_labels}
\begin{figure}[h!]
\centering
\includegraphics[width=\linewidth]{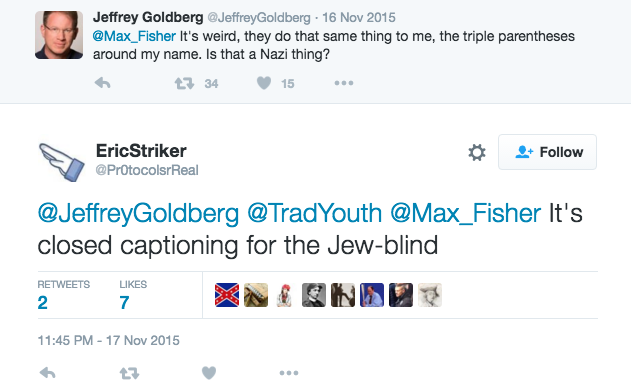}
\caption{A hate-monger responds to Jefferey Goldberg, Editor-in-Chief of The Atlantic, explaining the meaning of the echo. Note the Nazi salute used as the profile picture of the HM, and the user name -- a reference to the antisemitic conspiracy trope of the Protocols of the Elders of Zion.}
\label{fig:revelation}
\end{figure}

Clustering results were reported in Section \ref{sec:unsupervised}. While node colors in Figure \ref{subfig:mention_pred} are decided by cluster assignment, similar results are obtained when node colors are decided by the multi-modal neural architecture (Section \ref{sec:nn}). 
Most singletons are neutral users, in line with the trend presented in Figure \ref{subfig:mention_gold_sing}.
Hate-mongers, on the other hand, make the bulk of the large component and more likely be part of a connected component. This tendency is striking as hate-mongers tend to have significantly less friends and followers (see Table \ref{tab:class_stats}).
\rev{The discrepancy between their connectedness in the echo-induced network and their global degree suggests that the echo is more infectious as a hate-symbol/meme than in its other senses (a hug, broadcasting, etc.) -- highlighting the communal aspect in the adoption of hate. This is in line with previous work reporting that radical content travels faster and further in the network \cite{mathew2019spread}. These observations also provide a different angle on the notion of the `lone wolf' discussed in \cite{ribeiro2017like} -- on the one hand, hate mongers are highly active and organized, while on the other hand, their in and out degrees are significantly smaller than those of mainstream users.}  
It is interesting to see that some responders and neutral users are also at the core of the large components. We manually examined some of them, observing that responders often attract response from the hate-mongers. A typical exchange is presented in Figure \ref{fig:revelation}.

\begin{table}[t!]
\centering
\begin{tabular}{l|cccc}
User name & Suspended & Predicted  & Manual Label \\
\hline \hline
ThaRightStuff  & \cmark & HM  & \cmark \\
ramzpaul & \xmark & HM & \cmark \\
PaulTown & \cmark & HM & \cmark \\
Third\_Position & \crossedcmark & HM & \cmark \\
DrDavidDuke  & \xmark & HM & \cmark \\
SeventhSonTRS  & \cmark & HM & \cmark \\
TrumpHat & \cmark & HM & \cmark \\
TheeCurrentYear  & \cmark & HM & \cmark \\
\end{tabular}
\caption{Accounts with the highest generalized centrality. The not check mark (\crossedcmark) in the suspended column refers to accounts that are temporarily suspended (at the time of the query) as it violated the Twitter Media Policy.}
\label{table:high_centrality_users}
\end{table}

\subsection{Hate Leaders} 
Using network centrality measures, we can find leaders and promoters of racism and hate. Using the three network semantics (reply, mention and retweet) and seven centrality measures (in/out/total degree, betweenness, eigenvector, closeness and page-rank) we rank users by the number of times they appear in the $k$-core of most central users. We call this ranking {\em generalized centrality}. A number of users with the highest generalized centrality are listed in Table \ref{table:high_centrality_users}, some of which are known leaders of white supremacy movements, e.g., David Duke ({\em @DrDavidDuke}), the former Grand Wizard of the KKK, and Mike Enoch ({\em @TheRightStuff}), founder of The Right Stuff media network and The Daily Shoah podcast. The appearance of these notorious figures serves as a sanity check for the generalized centrality measure. To further verify the role of the central nodes we observe their label as predicted by the neural network model. As can be seen in Table \ref{table:high_centrality_users} all of the users presented are predicted as HM. Moreover, as a proxy, we queried Twitter for their account, finding that all of them were suspended -- an indication for high profile malicious, often racist, activity. We note that no N or R users were found to have high generalized centrality rank. 

The absence of other known leaders of the alt-right, e.g., Mike Cernovich and Richard Spencer, is somewhat surprising. We attribute it to the constraints imposed in the curation of the raw echo corpus (see Section \ref{sec:data}). These users may have not used the echo (in the 10\% sample) during the two month span the base corpus was curated. However, their centrality to the network is evident by the large number of times they are mentioned or being retweeted by the nodes flagged as hate-mongers, compared to the minimal trace they leave in the users of the R and N groups. Spencer, for example, is mentioned 5565 times, retweeted 1716 times, and replied to 1611 times by the HM group. These numbers are comparable to the mention/retweet/reply counts of the users with the top generalized centrality. \rev{These findings suggest that we managed to accurately reconstruct the network of hate mongers, in spite of the limitations and constraints imposed by Twitter API and other access issues.} 

\subsection{Linguistic Variations: Orthography and Semantics}
\label{subsec:linguistic}
\rev{We observe variations in the orthography and the semantics of the echo symbol. These variations are typically the result of the canonization of a word or a term within a certain speaker community, hence providing another perspective on adaptation of linguistic forms by a wider community. 
\paragraph{Abstraction and Semantic Drift} Starting as an abstract symbol, the echo was used to mark concrete named entities -- \emph{people} of Jewish heritage. It further evolved to mark \emph{abstract entities} such as \texttt{(((bankers)))} and \texttt{(((globalists)))}, echoing ancient antisemitic tropes. The use of the echo to mark abstract \emph{concepts} such as \texttt{(((narrative)))} or suggestive pronouns like \texttt{(((they)))} and \texttt{(((who)))} reflects another stage in the semantic evolution of the symbol. Finally, anecdotal evidence demonstrate that the antisemitic symbol is being repurposed to target other minority groups, e.g., \texttt{(((illegal mexicans)))}, \cite{tuters2019they}.}

\paragraph{Expressive Lengthening} \rev{Expressive lengthening, common in online informal writing, is the habit of adding characters to words in order to enhance the message or the sentiment conveyed in it. Typical examples are `aaaaaaaaaargh', `lolllll', and `sweeeeet!!!!' \cite{mcculloch2019because}. We observe expressive lengthening of the echo as hate-mongers try to underscore their hate, e.g., \texttt{((((((bankers))))))} and   \texttt{(((((jooooooos\footnote{A derogatory term for Jews, used for its (expressively-lengthened) homophony.})))))}. Another interesting orthographic phenomena is the use of a reverse echo to declare an `Aryan' affiliation, e.g., users declaring themselves \texttt{))))goyim\_godess((((} and \texttt{)))anti-semitic(((}.}

\subsection{The Intersectionality of Hate}
\label{subsec:ideas_hashtags}
\rev{When used as a hate-symbol, the echo is mostly used in an antisemitic context (see previous subsection for exceptions). Although one would expect a dataset constructed around the echo to contain mostly antisemitic hate speech, we do observe the ``intersectionality of hate'' which allows us to explore the attitude of hate groups toward other minorities and protected groups.}

Looking at the words and hashtags used most frequently by each of the groups, we observe a general racial pattern, going well beyond the antisemitic use of the echo. Users flagged as hate-mongers by our algorithm, are twenty times more likely to use the term Zionist as a general slur; talk about whiteness and white genocide; use derogatory terms like kike\footnote{Ethnic slur for Jews.}, cuck\footnote{A weak and submissive person. Similar to the classic `pussy'. Often used to describe minorities and ``intellectuals''.}, and skittle\footnote{Originally a small fruit-flavoured candy, repurposed as a derogatory term.}, referring to Arabs. 
\rev{In addition, these users are more likely to refer to Arabs, Muslims and immigrants in more explicit derogatory ways. For example, Muslims are addressed as muzzies\footnote{A religious slur referring to Muslims.} and the hashtag \texttt{\#rapefugees} is used to depict refugees as rapists.} 

\rev{The following tweet, posted by an HM account provides an illuminating example for the ``intersectionality'' of hate: {\footnotesize \texttt{Poland refuses \#rapefugees and is now on the verge of civil war. (((Who))) could be behind this? \#WhiteGenocide}}. Notice the dual strand of hate in that tweet: labeling Muslim refugees arriving in Europe as rapists, and the abstract use of the echo to hint that the influx of the ``rapefugees'' is a Jewish conspiracy to destabilize western countries as part of a war on the ``white race''.}

Comparing the popular hashtags among the HM and N groups we find that the HM group is trending with \#pizzagate, \#minorityPrivilige, \#WhiteGenocide, \#altright, \#tcot\footnote{A reference to the ``Top Conservatives On Twitter''.}, \#AmericaFirst, \#GamerGate, \#FeelTheBern, \#MAGA \#Brexit and \#rapefugees, while  the N echo users tend to use the hashtags \#job, \#sex, \#LIVE, \#broadcasting, \#party and \#NowPlaying, all associated with other meanings of the echo symbol -- a visual resemblance of an engulfing hug or a radio tower. 

It is interesting to note that the R users seem to exhibit a stronger interest in politics, compared to their N counterparts. The most frequently used hashtags of the R group are: \#localbuzz, \#Facebook, \#SocialMedia, \#antisemitism, \#DemDebate, \#VPDebate, \#Israel, \#LonelyConservative, and \#NeverTrump, as well as \#MAGA and \#Trump2016. \rev{We attribute this tendency to an inherent selection bias -- the responders are those who care more about political agenda and therefore try to have a better grasp of its extreme fringes.}
These findings also support the split of the non hate-mongers users to two different clusters with unique features instead of a single large cluster that combines both groups.

We wish to stress that while the HM users of the echo tend to enthusiastically support a separatist right-wing agenda, not all conservative users or supporters of Trump or of the Brexit are hate-mongers. We also wish to point out that the perceived nicety of the R+N users, demonstrated by the heavy use of positive words is somewhat misleading. This may be a side-effect caused by the manner in which the corpus was constructed, since the meanings and the contexts in which the echo symbol is used are polarized. While the vast majority of the tweets in the corpus do not contain the echo at all, all users in the data did use this unique symbol, often as a very strong sentiment/stance marker. 

\subsection{Links to Russian Trolls} 
Recent studies suggest that foreign activity on social media was strategically used in an attempt to further radicalize groups that already have an inclination to extremism \cite{jamieson2018cyberwar,addawood2019linguistic}. We conclude this paper with a brief examination of foreign involvement with alt-right communities.

The Internet Research Agency (IRA) is a Russian troll-farm linked to the Russian intelligence, according to a declassified report by the United States Office of the Director of National Intelligence \shortcite{office2017assessing}, and the Special Counsel report on the Investigation into Russian Interference \cite{mueller2019report}. A list of 3,814 account handles, linked to the IRA was identified and released by Twitter. In the ten-week period preceding the 2016 election these accounts posted 175,993 Tweets, approximately 8.4\% of which were election-related \cite{twitter2018ira}. None of the IRA trolls used the echo in the \%10 sample of two months covered in the base corpus. However, we find their impressions in the network. Analysis of the data reveals that hate-mongers (HM) are eight to nine times more likely to mention or retweet an IRA user than their R+N counterparts. Looking only at users that actively engage with IRA accounts, a hate-monger engages with an IRA account in a higher rate, see Table \ref{tab:ira} for more details. While a detailed analysis of these efforts are beyond the scope of this paper, our computational results support the qualitative analysis of foreign meddling in local politics \cite{jamieson2018cyberwar}.

\begin{table}[]
    \centering
    \begin{tabular}{l|cc||cc}
    & \multicolumn{2}{c||}{GOLD USRS} & \multicolumn{2}{c}{ALL USRS} \\
    \hline
    Label & HM & R+N & HM & R+N \\
    \hline
    \#Users  & 170 & 830 & 1136 & 5937 \\
    \hline
    \#Users mentioning IRA & 88 & 67 & 623 & 379 \\
    \#User retweeting IRA & 81 & 53 & 529 & 312 \\
    \#Unique IRA mentioned & 24 & 25 & 63 & 76 \\
    \#Unique IRA retweeted & 19 & 20 & 45 & 46 \\
    \#Total IRA mentions & 196 & 102 & 1375 & 595 \\
    \#Total IRA retweets & 167 & 79 & 1088 & 479 \\
    \end{tabular}
    \caption{Engagement of echo users with IRA accounts. \#Users mentioning/retweeting IRA: the number of echo users mentioning/retweeting an IRA user. \#Unique IRA mentioned/retweeted: the number of unique IRA accounts mentioned by echo users. \#Total IRA mentions/retweets: the total number of mentions/retweets of IRA users by echo users.}
    \label{tab:ira}
\end{table}

\section{Conclusion}
\label{sec:conclusions}
Antisemitism is only one manifestation of racism. Using a large and unique corpus constructed around an ambiguous antisemitic meme, we showed how networks of hate-mongers can be reconstructed. Analyzing content and the network structure in tandem provides significant insights on the promotion of hate, beyond antisemitism, the central figures dominating the network, the engagement between hate-mongers and other users and the utilization of this network for international political warfare. Future work includes a temporal analysis of the formation of the network as well as a finer analysis of the types of hate promoted by the network.  

\bibliography{echo}
\bibliographystyle{aaai}
\end{document}